\documentclass[runningheads]{llncs}
\usepackage{graphicx}
\usepackage{hyperref}
\usepackage{multirow}
\usepackage{amsmath}

\begin{document}
\title{Self-Supervised Approach to Addressing Zero-Shot Learning Problem}

\author{Ademola Okerinde \and Sam Hoggatt \and 
Divya Vani Lakkireddy \and Nolan Brubaker \and William Hsu \and Lior Shamir \and Brian Spiesman}

\institute{Kansas State University, 2164 Engineering Hall, Manhattan, KS 43017-6221, USA  \\
\email{\{okerinde, schoggatt, divyalakki, ncbrubaker,\\ bhsu, lshamir, bspiesman\}@ksu.edu}}




\maketitle 

\begin{abstract}
In recent years, self-supervised learning has had significant success in applications involving computer vision and natural language processing. The type of pretext task is important to this boost in performance. One common pretext task is the measure of similarity and dissimilarity between pairs of images. In this scenario, the two images that make up the negative pair are visibly different to humans. However, in entomology, species are nearly indistinguishable and thus hard to differentiate. In this study, we explored the performance of a Siamese neural network using contrastive loss by learning to push apart embeddings of bumblebee species pair that are dissimilar, and pull together similar embeddings. Our experimental results show a 61\% F1-score on zero-shot instances, a performance showing 11\% improvement on samples of classes that share intersections with the training set.

\end{abstract}




\section{Introduction}
\label{introduction}

Zero-shot learning (ZSL) is a problem in machine learning, where during test time, a learner observes samples from classes not observed during training when the learner needs to predict sample class \cite{enwiki:1018108538}.


Deep learning (DL) requires large amounts of carefully labeled data, which is very often difficult to acquire and expensive to annotate. Even with large amounts of data, supervised learning still has blind spots in learning useful and rich representations. Moreover, the supervision signal can bias the network in unexpected ways. Self-supervised learning (SSL) is an effective way to extract training signals from massive amounts of unlabeled data and to learn good representation to facilitate downstream tasks where it is expensive to collect task-specific labels. This representation is a feature vector that contains semantic information about the input image. SSL is a special type of representation learning that enables learning good data representation from unlabeled dataset. SSL is necessary because data labeling is often expensive and thus high-quality labeled datasets are limited. Also, a good learned representation enables easier transfer of useful information/signals to a variety of downstream tasks. In our research, the downstream tasks consists of zero-shot and few-shot transfer to new tasks.

The use of contrastive loss as the cost function is critical to the success of SSL. This objective function pulls together embeddings of augmentations of the same bumblebee species pair and pushes embeddings of views of different species pairs far apart. This is arguably a challenging task, even to humans, because many species of bumblebees are quite similar, even to experts.

\begin{figure}
  \centering
  \includegraphics[width=\linewidth]{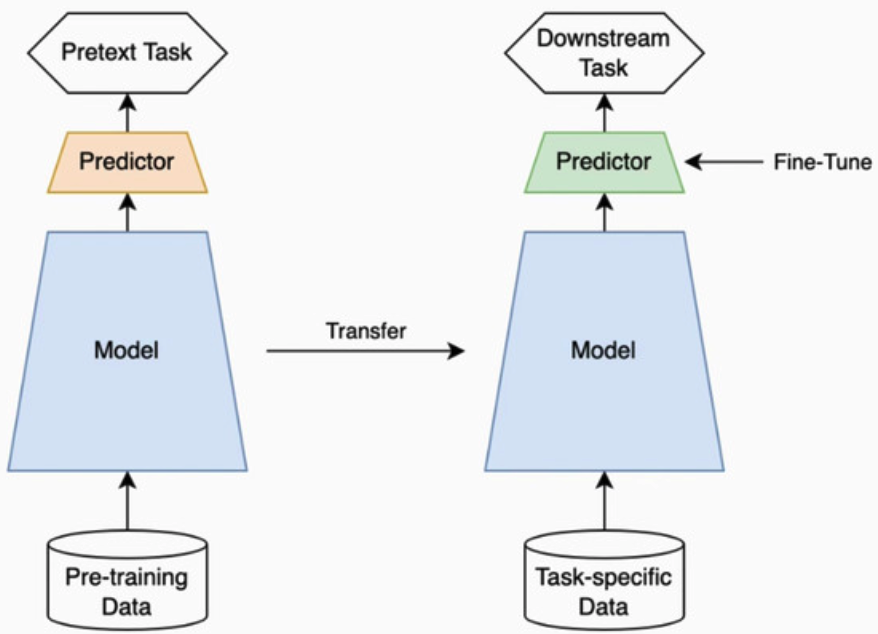}
  \caption{Pretext Tasks}
  \label{fig:pretext_tasks}
\end{figure}

Pretext tasks shown in Figure~\ref{fig:pretext_tasks} are an important component of any SSL training pipeline. Our goal was to have a model learn on downstream tasks like ZSL although models trained on pretext tasks were transferred for specific tasks. On some occasions, pre-training data could be the same as the task-specific data. In our research, we explored two options/approaches. Our pre-training data could be either bumblebees images or a CIFAR-10 dataset. In recent years, researchers have become more creative at determining different forms of pretext tasks. In \cite{gidaris2018unsupervised}, the task was to have the model predict the degree of rotation in an image. After a patch, \cite{doersch2015unsupervised} predicted which of eight neighboring locations had another patch. This forces the model to learn the innate relationships among patches of the input image. In our research, we formulated our pretext task as a measure of similarity in the same species and dissimilarity between different species of bumblebees.

Existing image-text retrieval models are often fine-tuned on pretext tasks in which the data distribution shift is minimal, but the domain of entomology has terms that are rare in most pre-trained tasks.

The Siamese neural network (SNN) is an architecture that comprises two identical neural networks with shared weights; their parameters are trained to determine the similarity between two data samples using a distance metric on the embeddings. Our research used SNN to determine if two bumblebee images belong to the same specie.

\section{Related Work}
\label{related_work}

Recently, SSL has gained a lot of momentum as a new paradigm of representation learning because supervised learning often requires a large number of labels, which is time consuming or not available in certain domains. As a result, researchers have become more creative: using part of the data itself as a form of supervision, a seemingly frivolous exercise that has produced remarkable outcomes.

Self-supervision is often achieved either through self-prediction or contrastive learning (CL).
In \cite{vondrick2018tracking}, video colorization as a self-supervised learning method provides a rich representation that can be used for video segmentation and unlabeled visual region tracking without extra fine-tuning. Without using supervised labels for training, the zero-shot CLIP \cite{radford2021learning} classifier performs quite well on challenging image-to-text classification tasks.
\cite{jia2021scaling} learned a joint visual-language representation using CL and training on one billion noisy image alt-text pairs. To the best of our knowledge, such an enormous dataset is not yet available in entomology, which makes the learning task harder.

CL has helped make many SSL applications successful. Ideally, CL keeps positive pairs close to each other and negative pairs far apart in the feature space. According to \cite{chen2020simple,grill2020bootstrap}, CL currently performs at state-of-the-art level in SSL. \cite{wu2018unsupervised} adopted a memory bank to store the instance class representation vector. Momentum Contrast (MoCo) \cite{he2020momentum} trained a visual representation encoder by matching an encoded query q to a dictionary of encoded keys with contrastive loss. SimCLR \cite{chen2020simple} used the normalized temperature-scaled cross-entropy loss as contrastive loss. BYOL \cite{grill2020bootstrap}  relied only on positive pairs, not negative pairs, to learn feature representation. SimSiam \cite{chen2020simple,he2020momentum} removed the momentum encoder and provided a simpler design based on BYOL. 

In the first paper in which SNN was published, \cite{bromley1993signature}, the author depicted the significance of the network for the signature verification task. The goal was to identify signature authenticity.

ZSL is a model's ability to detect classes never seen during training. Even for humans, this is difficult. These classes are sometimes referred to as Out Of Distribution (OOD) samples. Earlier work in ZSL use attributes in a two-step approach to infer unknown classes. In vision research, more recent, advanced models learn mappings from image feature space to semantic space. \cite{narayan2020latent} proposed a TF-VAEGAN to address generalized zero-shot recognition. In that research, both visual features and corresponding semantics are fed into the architecture. In contrast, we based our approach on only visual signals.

\section{Proposed Approach}
\label{approach}
\begin{figure}
  \centering
  \includegraphics[width=\linewidth]{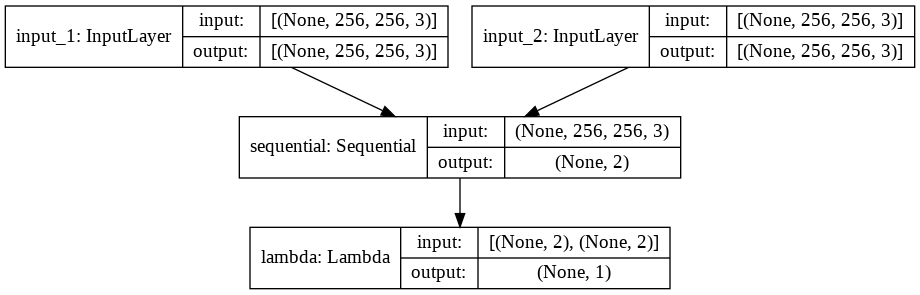}
  \caption{Siamese Neural Network}
  \label{fig:siamese_network}
\end{figure}

\begin{figure}
    \centering
    \includegraphics[width=0.3\textwidth]{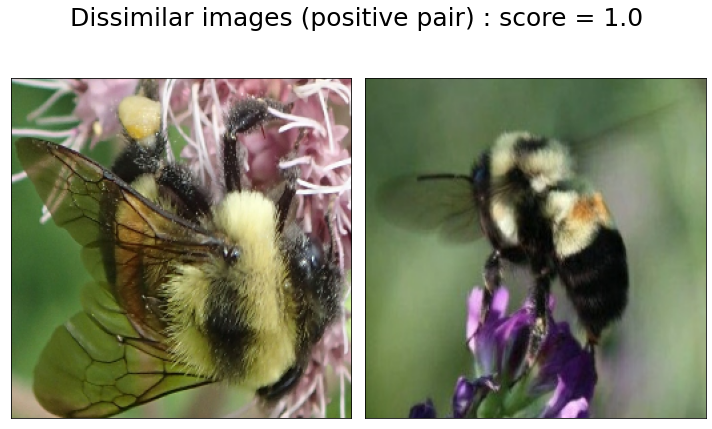}
    \includegraphics[width=0.3\textwidth]{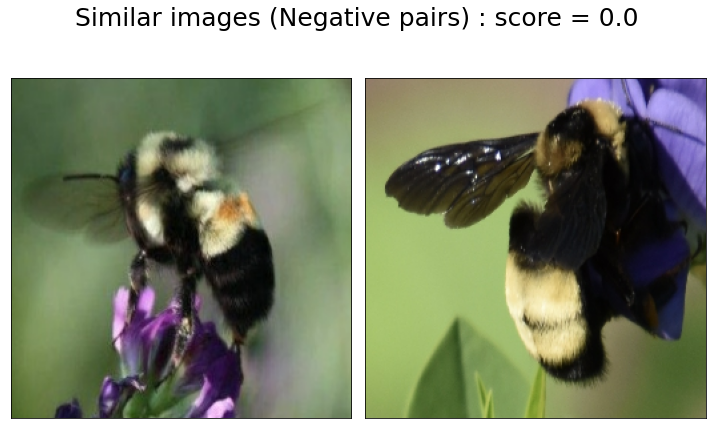}
    \caption{Positive pair and Negative pair}
    \label{fig:positive_and_negative_pair}
\end{figure}

To determine how well our model used a pretext task in a downstream task of novelty detection, we trained our model not on an output  class of bumblebees but rather
the similarities and dissimilarities between differrent species using contrastive loss as shown in Figure~\ref{fig:positive_and_negative_pair}. This task forces the model to be aware of visual signals needed in the downstream task. Figure~\ref{fig:siamese_network} shows our convolutional SNN, which uses feature extractors pretrained on ImageNet.

The SNN has three convolutional layers with 16, 32, and 64 filters. We use 3*3 kernel size with the same padding. The final fully connected layer has 128 neurons on which the energy function is computed to estimate the distance vector. Euclidean distance is used as the energy function. Rectified linear activation function (ReLU) is applied after each conv layer followed by maxpooling2D. To avoid overfitting, we applied dropout with probability of 20\% after the feature extraction stage. Before passing the image pairs to the feature extractors, we augmented the images by randomly flipping and rotating.
We use contrastive loss as the objective function to be optimized. The loss equation is defined in Equation \ref{loss_eqn}

\begin{equation} 
\begin{split}
L(Y,B1,B2,W) = (1-Y) \cdot \frac{1}{2}D_w^2 + Y\frac{1}{2}(max(0,m-D_w))^2 \\
\\
D^2_w = (f(B1) - f(B2))^2
\end{split}
\label{loss_eqn}
\end{equation}
$B1, B2$ are bumblebee specie image pair.
 $D_w^2$    is the euclidean distance which represent the energy function. $Y$ can either be 1 or 0 depending or whether the pair is similar or not. And, $m$ represent margin, a threshold for which the embedding is satisfactory separable. $W$ are the network parameters to be learnt during training.

We experimented with MobileNetV2 and ResNet-101 as feature extractors before measuring the energy function between each embedding of the input image pair.


\section{Methodology and Experimental Design}

\subsection{Dataset}
\label{Dataset}

\begin{figure}
  \centering
  \includegraphics[width=\linewidth]{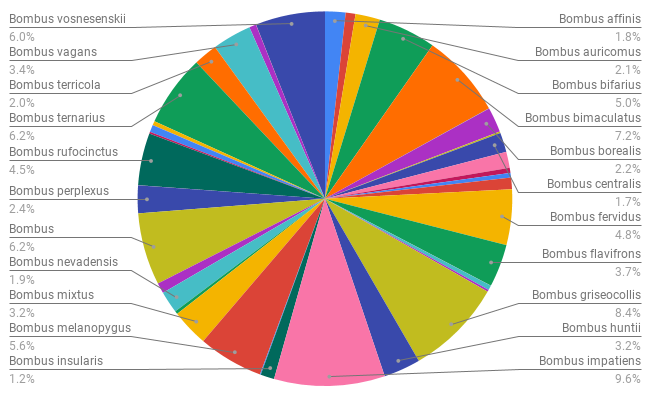}
  \caption{Pie chart of the various species}
  \label{fig:histogram}
\end{figure}

The dataset consisted of 45 species of bumblebee totalling 104770 samples as shown in Table~\ref{tab:dataset}.  Figure~\ref{fig:histogram} shows the percentage of the different species in an histogram. Bombus impatiens dominates with 9.6\% of the entire dataset.


\begin{table}[ht]
  \caption{Bumblebee (B.) Species. (\textbf{Zero and few shots are in bold})}
  \label{tab:bumble_bee_species}
  \begin{tabular}{lc|lc}
  \hline
Species  & \#samples & Species &  \#samples  \\
\hline
 \textbf{B. appositus} & \textbf{888} & B. affinis & 1855  \\
 \textbf{B. crotchii} & \textbf{457} & B. bifarius & 5237 \\
 \textbf{B. bohemicus} & \textbf{21} & B. bimaculatus & 7595  \\
 \textbf{B. caliginosus} & \textbf{147} & B. borealis & 2256 \\
 \textbf{B. franklini} & \textbf{9} & B. citrinus & 1449\\
 \textbf{B. cockerelli} & \textbf{3} & B. centralis & 1755   \\
 \textbf{B. cryptarum} & \textbf{419} & B. fernaldae flavidus & 1047\\
 \textbf{B. suckleyi} & \textbf{14} & B. flavifrons & 3866\\
 \textbf{B. fraternus} & \textbf{424} & B. fervidus californicus & 5020   \\
 \textbf{B. frigidus} & \textbf{179} & B. griseocollis & 8787\\
 \textbf{B. variabilis} & \textbf{12} & B. impatiens & 10008\\
 \textbf{B. jonellus} & \textbf{5} & B. insularis & 1241  \\
 \textbf{B. kirbiellus} & \textbf{69} & B. melanopygus & 5892\\
 \textbf{B. morrisoni} & \textbf{276} & B. mixtus & 3359  \\
 \textbf{B. natvigi} & \textbf{3} & B. nevadensis & 1982\\
 \textbf{B. occidentalis} & \textbf{943} & B. pensylvanicus sonorus & 6522\\
 \textbf{B. polaris} & \textbf{11} & B. perplexus & 2503 \\
 \textbf{B. sandersoni} & \textbf{163} & B. rufocinctus & 4719 \\
 \textbf{B. sitkensis} & \textbf{647} & B. huntii & 3335 \\
 \textbf{B. sylvicola} & \textbf{366} & B. ternarius & 6467\\
 \textbf{B. vandykei} & \textbf{625} & B. terricola & 2145 \\
 & & B. vagans & 3546 \\
 & & B. vosnesenskii & 6284 \\
 & & B. auricomus & 2219  \\
\hline
\hline
& & Total =  104770 \\
  \hline
  \end{tabular}
\end{table}


\begin{table}[hbt]
\centering
  \caption{ResNet-101 confusion matrix at 0.5 threshold on testset-ALL (45 species)}
  \label{tab:confusion_matrix_resnet_testset_all}
  \begin{tabular}{l|c|c}
& Predicted similar & Predicted dissimilar \\
\hline
Actually similar & 61 & 29  \\
Actually dissimilar & 43 & 47  \\
  \hline
  \end{tabular}
\end{table}

\begin{table}[hbt]
\centering
  \caption{ResNet-101 confusion matrix at 0.5 threshold on testset (24 species)}
  \label{tab:confusion_matrix_resnet_testset_24_species}
  \begin{tabular}{l|c|c}
& Predicted similar & Predicted dissimilar \\
\hline
Actually similar & 3725 & 1267  \\
Actually dissimilar & 1514 & 3478  \\
  \hline
  \end{tabular}
\end{table}

\begin{table}[hbt]
\centering
  \caption{ResNet-101 Confusion matrix at 0.5 threshold on 21 unknown test set species - zero-shot learning}
  \label{tab:confusion_matrix_resnet_testset_zsl}
  \begin{tabular}{l|c|c}
& Predicted similar & Predicted dissimilar \\
\hline
Actually similar & 24 & 18  \\
Actually dissimilar & 15 & 27  \\
  \hline
  \end{tabular}
\end{table}

\begin{table}[hbt]
\centering
  \caption{F1-scores of zero-shot specie pair}
  \label{tab:species_comparison}
  \begin{tabular}{l|c|c|c|c|c}
& B. natvigi & B. variabilis & B. appositus & B. fraternus & B. frigidus\\
\hline
B. cockerelli & 0.5  & \textbf{0.88}  & 0.28 & \textbf{1.00} & \textbf{1.00}\\
B. polaris & 0.5 &  0.43 & 0.50 & 0.72 & 0.46\\
B. occidentalis & 0.11 &  0.48 & 0.52 & 0.60 & 0.63 \\
B. cryptarum & 0.62 & 0.49 & 0.51 & \textbf{0.74} & 0.58 \\
B. caliginosus & 0.50 & 0.39 & 0.52 & 0.60 & 0.59 \\
  \hline
  \end{tabular}
\end{table}

\begin{table}[hbt]
\centering
  \caption{ResNet-101 confusion matrix at 0.5 threshold on validation}
  \label{tab:confusion_matrix_resnet}
  \begin{tabular}{l|c|c}
& Predicted similar & Predicted dissimilar \\
\hline
Actually similar & 3197 & 1123  \\
Actually dissimilar & 1243 & 3077  \\
  \hline
  \end{tabular}
\end{table}

\begin{table}[ht]
\centering
  \caption{MobileNetV2 confusion matrix at 0.5 threshold on validation}
  \label{tab:confusion_matrix}
  \begin{tabular}{l|c|c}
& Predicted similar & Predicted dissimilar \\
\hline
Actually similar & 2897 & 1423  \\
Actually dissimilar & 1608 & 2712  \\
  \hline
  \end{tabular}
\end{table}


\begin{table}[ht]
\centering
  \caption{Number of samples in the dataset.}
  \label{tab:dataset}
  \begin{tabular}{lcc}
  \hline
Dataset  & images & species-represented \\
\hline
Train + Validation  & 79280 & 24\\
Test  & 25490 & 45 \\
  \hline
  \end{tabular}
\end{table}

\subsection{Data Preprocessing}

We preprocessed the dataset by moving all the species with fewer than a thousand samples to the test set, which means that 21 species are not represented in the training set. We then randomly selected 20\% of samples of each remaining species to include in the test data. All 45 species are represented in the test data. 

\begin{figure}
    \centering
    \includegraphics[width=0.4\textwidth]{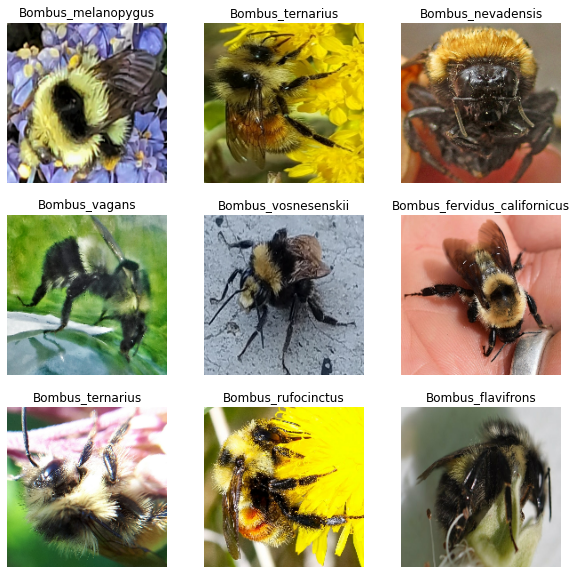}
    \caption{Example images of various bee species}
    \label{fig:dl_sl}
\end{figure}

\subsection{Experimental results}

We trained a  convolutional SNN for a maximum of 100 epochs, with patience level of 7, on 80\% of the training set. We then evaluated the trained model on the 20\% validation set. The results are shown in Table~\ref{tab:experiments}.

\section{Results and Discussion}

Table~\ref{tab:experiments} shows the average accuracy, recall, precision, and F1 score at different thresholds/cut-offs. Of all the thresholds, the results reveal that 0.5 is a good decision boundary for most species. Similarity scores below the threshold are classified as similar and vice versa. We also inspected the performance of contrastive learning within each species of the bumblebees to discover hard and easy data pairs. On the test data, the best performing SNN model, which uses ResNet-101 as base feature extractor, achieves 73\% precision, 72\% recall, 73\% F1-score, and 72\% accuracy.

The results shown in Table~\ref{tab:results_testset} show that our SSL network can detect zero shot instances of the bumblebees although the SNN performs much better on species that share intersection with the training dataset. Table~\ref{tab:species_comparison} shows that our network can still differentiate between pairs of zero-shot instances. The network could differentiate the {\it B. cockerelli} and {\it B. fraternus} pair as well as the {\it B. cryptarum} and {\it B. fraternus} pair with a high confidence. This is useful in deciding in what species the novel instance falls. With this, an entomologist can quickly narrow down a novel species class to either in-distribution or OOD samples.

\subsection{Conventional Zero-shot learning}

This involves learning a classifier that can detect only unknown test samples. Our network achieved 61\% F1-score on 5681 unknown species of bumblebees. As explained in \cite{okerinde2021egan}, this task is made more difficult because of a high class imbalance with some species having very few samples, as can be seen in Table~\ref{tab:bumble_bee_species}.

\subsection{Generalized Zero-shot learning}

The goal of generalized zero-shot learning is to train a classifier that can detect test instances from seen or unseen classes. This is more useful, and thus more challenging. With the entire 45 species in the test set, our network achieved a 60\% F1-score on various pairing of the species as shown in Table~\ref{tab:results_testset}.


\section{Conclusion}

In this work, we explored using a self-supervised learning approach to detect novel bumblebee instances in a zero-shot setting. We showed that our model performs well in learning the embeddings of similar and dissimilar bumblebee species pairs. Our experimental results showed that our model can generalize to zero shot instances that were not part of the training. Future work will focus on improving the performance on the ZSL task by learning a multi-modal network that maps images to text-attribute descriptions. We can also explore cosine similarity as the energy and triplet loss as the loss function.

\begin{table}[ht]
\centering
  \caption{Classification Accuracy, F1, Precision, Recall when using different methods on validation set.}
  \label{tab:experiments}
  \begin{tabular}{lcccccc}
  \hline
Models & Threshold & Precision & Recall & F1 & Accuracy\\
\hline
\multirow{1}{10em}{SNN + ResNet-101\cite{he2016deep}}
& 0.5  & 0.73 & 0.72 & 0.73 & 0.72 \\
\hline

\multirow{9}{4em}{SNN + MobileNetV2\cite{sandler2018mobilenetv2}} & 0.1 & 0.66 & 0.51 & 0.35 & 0.50 \\
& 0.2 & 0.62 & 0.55 & 0.47 & 0.55 \\
& 0.3 & 0.63 & 0.60 & 0.58 & 0.60 \\
& 0.4 & 0.63 & 0.63 & 0.64 & 0.63 \\
& 0.5 & 0.65 & 0.65 & 0.65 & 0.65\\
& 0.6 & 0.65 & 0.65 & 0.64 & 0.65 \\
& 0.7 & 0.67 & 0.66 & 0.65 & 0.66 \\
& 0.8 & 0.69 & 0.65 & 0.63 & 0.65 \\
& 0.9 & 0.69 & 0.62 & 0.58 & 0.62 \\
\hline
\hline

  \end{tabular}
\end{table}

\begin{table}[ht]
\centering
  \caption{Classification Accuracy, F1, Precision, Recall on test set using ResNet-101 as feature extractor.}
  \label{tab:results_testset}
  \begin{tabular}{lcccccc}
  \hline
 & Threshold & Precision & Recall & F1 & Accuracy\\
\hline
21 species (Zero-Shot) & 0.5  & 0.61 & 0.61 & 0.61 & 0.61 \\
45 species (ALL) & 0.5  & 0.61 & 0.60 & 0.60 & 0.60 \\
24 species & 0.5  & 0.73 & 0.72 & 0.72 & 0.72 \\

\hline

  \end{tabular}
\end{table}

 


\bibliographystyle{splncs03_unsrt}
\bibliography{references}

\end{document}